\documentclass{article}
\usepackage{spconf,amsmath,amssymb,graphicx,booktabs,multirow,hyperref}

\newcommand{\rulehoef}{\textsc{Hoeffding}}
\newcommand{\rulecp}{\textsc{CP}}
\newcommand{\rulewsr}{\textsc{WSR}}
\newcommand{\rulenaive}{\textsc{Naive}}

\title{False sense of safety in selective signal classification:\\
auditing bound tightness and exchangeability for risk control}

\name{Jingwen Zhou, Mingzhe Wang}
\address{Xidian University, Xi'an, China}

\begin{document}
\ninept
\maketitle

\begin{abstract}
Selective prediction with distribution-free risk control promises that, with
confidence $1-\delta$ over the calibration draw, the error rate of accepted
inputs stays below a user budget $\alpha$. We audit this promise on
signal-domain detectors---machine anomalous-sound detection (ASD) and
AI-generated-image forensics---for four calibration rules: uncertified
empirical thresholding (\rulenaive{}) and certified Hoeffding,
Clopper--Pearson (\rulecp{}), and betting (\rulewsr{}) upper confidence
bounds. We report three findings. (i) \rulenaive{} thresholding, common in
practice, exceeds its declared budget in 49--73\% of synthetic trials
($n{=}200$ calibration points) and in up to 68\% of real-data splits: a false
sense of safety rather than a broken theorem, since the rule never had a
certificate. (ii) Tightness matters: \rulecp{} and \rulewsr{} certify
substantial coverage where Hoeffding certifies none, with zero observed
budget overruns under exchangeable splits. (iii) Under grouped deployment
(unseen machine types or generators), certified rules overrun in 9--30\% of
trials---far above $\delta$---showing the failure lies in the broken
exchangeability premise, not in the bounds; a conservative per-group
threshold restores validity at a severe coverage cost. 
\end{abstract}


\section{Introduction}
\label{sec:intro}
Signal-domain detectors are rarely perfect, and many applications would
rather abstain than err: an anomalous-sound-detection (ASD) system can defer
a machine to human inspection, and an image-forensics system can flag an
image as ``undecided''. Selective classification, from Chow's classical
reject option~\cite{chow1970} to its modern
treatments~\cite{elyaniv2010,geifman2017}, formalizes
this with an accept/abstain threshold on a confidence score.
Distribution-free risk control---risk-controlling
prediction sets (RCPS)~\cite{bates2021rcps}, Learn-then-Test
(LTT)~\cite{angelopoulos2021ltt}, and conformal risk
control~\cite{angelopoulos2024crc}, all rooted in conformal
prediction~\cite{vovk2005,angelopoulos2021gentle} and its split
variant~\cite{papadopoulos2002}---promises a finite-sample
certificate: with probability at least $1-\delta$ over the calibration draw,
the error rate among accepted inputs (the \emph{selective risk}) is at most a
user budget $\alpha$. Such certificates now reach applications from
set-valued classification~\cite{sadinle2019} and biomedical
imaging~\cite{angelopoulos2022image} to language-model
factuality~\cite{mohri2024}, making their reliability on deployed signal
systems a timely question.

In practice, two failure modes undermine this promise and are routinely
ignored in signal-domain deployments. First, practitioners often tune the
threshold so that the \emph{empirical} calibration risk is below $\alpha$
and report the system as ``risk-controlled''. This rule---we call it
\rulenaive{}---carries no certificate, so its failures break no theorem;
the harm is the \emph{false sense of safety} created when its declared
budget is read as a guarantee, much as miscalibrated confidence scores
mislead downstream decisions~\cite{guo2017}. How often, and by how much, the
budget is exceeded at test time has not been quantified on signal data.
Second, all certificates assume calibration and test data are exchangeable,
but realistic signal-domain splits are \emph{grouped}: the deployed system
meets machine types or generative models absent from calibration.
Weighted and adaptive conformal methods for covariate
shift~\cite{tibshirani2019,gibbs2021}, robust prediction sets under
$f$-divergence shifts~\cite{cauchois2024}, and conformal inference beyond
exchangeability~\cite{barber2023} all relax this premise for coverage
guarantees, but the magnitude
of the damage to selective-risk certificates on signal data is unknown.

Closest to our work, Basu~\cite{basu2026} ablates nine finite-sample bound
families (including Hoeffding, Clopper--Pearson, and WSR betting) for
selective prediction on NLP intent-classification benchmarks and proposes a
transfer-informed warm start for the betting bound. Our audit is
complementary and differs in three ways: we target signal domains (ASD,
image forensics) with their characteristic grouped deployment structure; we
quantify the false-safety behaviour of the uncertified \rulenaive{} rule,
which~\cite{basu2026} does not consider; and rather than assuming a benign
source domain, we isolate---via matched random-vs-grouped splits on the same
data---how much of the observed failure is attributable to broken
exchangeability rather than to the bound itself.

\textbf{Contributions.} (1)~A unified audit of four selective thresholds
(\rulenaive{}; \rulehoef{}~\cite{hoeffding1963};
Clopper--Pearson~\cite{clopper1934}; the Waudby-Smith--Ramdas betting bound,
\rulewsr{}~\cite{waudbysmith2024}) under one protocol on synthetic data, real
ASD scores (DCASE~2023 Task~2 development set, 7 machine types, BEATs
embeddings, 4 score backends)~\cite{dohi2023dcase,chen2023beats}, and real
forensics scores (GenImage, 7 generators)~\cite{zhu2023genimage}.
(2)~Quantitative evidence of the false sense of safety and of the coverage
recovered by tight bounds. (3)~A controlled exchangeability ablation showing
budget overruns under group shift are caused by the broken premise, not the
rules, plus a simple per-group conservative mitigation and its trade-off.

\section{Selective risk control: rules and bounds}
\label{sec:method}

\subsection{Setup and notation}
A frozen detector emits a scalar risk score $s(x)$, where larger means more
likely to be an error event $y{=}1$ (an anomaly missed, a fake accepted as
real, etc.). Given a threshold $\lambda$, the system accepts $x$ iff
$s(x)\le\lambda$. The \emph{selective risk} is
$R(\lambda)=\Pr\{y{=}1 \mid s(x)\le\lambda\}$ and the \emph{coverage} is
$C(\lambda)=\Pr\{s(x)\le\lambda\}$, i.e.\ the fraction of inputs the system
answers. Given calibration data $\{(s_i,y_i)\}_{i=1}^{n}$, budget $\alpha$,
and confidence parameter $\delta$, every rule picks the largest threshold
whose certified (or, for \rulenaive{}, empirical) risk is within budget:
\begin{equation}
\hat\lambda = \max\bigl\{\lambda\in\{s_{(k)}\}_{k=1}^{n} :
\mathrm{UCB}_{\delta/n}\bigl(\{y_i: s_i\le\lambda\}\bigr)\le\alpha\bigr\},
\label{eq:rule}
\end{equation}
where $s_{(1)}\le\dots\le s_{(n)}$ are the sorted calibration scores and
$\mathrm{UCB}_{\delta'}(\cdot)$ is an upper confidence bound at level
$\delta'$ on the mean of the indicated losses. The Bonferroni correction
$\delta/n$ over the $n$ candidate thresholds makes the certificate valid
simultaneously over the search, as in LTT~\cite{angelopoulos2021ltt}. If no
candidate passes, the system always abstains ($C{=}0$, trivially safe). We
call a test run a \emph{budget violation} if the empirical test selective
risk exceeds $\alpha$ with at least one accepted sample. For certified rules
this event has probability at most $\delta$ under exchangeability; for
\rulenaive{} no such promise exists, and its observed violation rate
quantifies the false sense of safety, not a broken proof.

\subsection{The four rules}
\begin{figure}[t]
\centering
\includegraphics[width=\columnwidth]{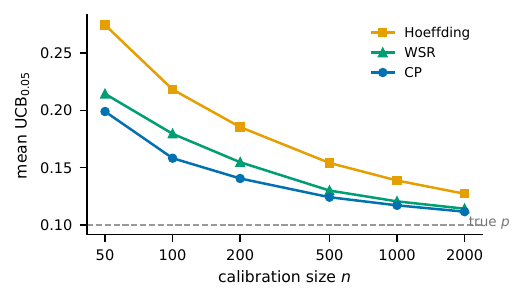}
\caption{Bound tightness: mean $95\%$ upper confidence bound on a Bernoulli
rate $p{=}0.1$ vs.\ calibration size $n$ (200 Monte-Carlo draws per point).
\rulecp{} is tightest for binary losses; \rulewsr{} tracks it closely and
\rulehoef{} lags, which explains its zero certified coverage at small
$\alpha$ in Table~\ref{tab:synth}.}
\label{fig:tightness}
\end{figure}

\textbf{\rulenaive{}:} $\mathrm{UCB}:=\hat p$, the empirical risk of the
accepted prefix. \quad
\textbf{\rulehoef{}:} $\mathrm{UCB}:=\hat p+\sqrt{\ln(1/\delta')/(2k)}$ for a
prefix of size $k$~\cite{hoeffding1963}; valid but loose at small $\alpha$.
\quad
(Empirical-Bernstein refinements~\cite{maurer2009} sharpen the constant but
not the qualitative picture below.) \quad
\textbf{\rulecp{}:} the exact binomial upper limit~\cite{clopper1934},
$\mathrm{Beta}^{-1}(1-\delta';x{+}1,k{-}x)$ for $x$ errors among $k$;
admissible for binary losses. \quad
\textbf{\rulewsr{}:} the betting/hedged bound of~\cite{waudbysmith2024},
an instance of the confidence-sequence and e-value machinery of
game-theoretic statistics~\cite{howard2021,ramdas2023}.
With running mean $\hat\mu_t=\tfrac{1/2+\sum_{i\le t}x_i}{t+1}$ and
variance $\hat\sigma_t^2=\tfrac{1/4+\sum_{i\le t}(x_i-\hat\mu_i)^2}{t+1}$,
form bets and wealth
\begin{equation}
\nu_t=\min\Bigl\{1,\sqrt{\tfrac{2\ln(1/\delta')}{n\hat\sigma_{t-1}^2}}\Bigr\},
\quad
K_t(R)=\prod_{i\le t}\bigl(1-\nu_i(x_i-R)\bigr).
\label{eq:wsr}
\end{equation}
$K_t(R)$ is a nonnegative supermartingale when $\mathbb{E}[x]\ge R$, so
$\mathrm{UCB}:=\inf\{R:\max_t K_t(R)>1/\delta'\}$, found by bisection since
the wealth is monotone in $R$. Three conservative implementation choices:
the bisection returns the rejected endpoint (upward bias); each candidate
prefix is fed to the martingale in a data-independent random order
(score-sorted order would break the martingale property); and we keep the
Bonferroni correction over candidate thresholds rather than exploiting
time-uniformity. A Monte-Carlo self-test ($p{=}0.1$, $n{=}200$,
$\delta{=}0.05$, 200 draws) gives mean upper bounds $0.187$ (\rulehoef{})
$>0.156$ (\rulewsr{}) $>0.143$ (\rulecp{}), with empirical coverage of the
\rulewsr{} bound $\ge 94.5\%$; Fig.~\ref{fig:tightness} traces the gap over
calibration sizes.

\textbf{When can a bound certify anything?} The zero-error prefix gives a
useful closed form. With $x{=}0$ errors among $k$ accepted calibration
points and $\delta'{=}\delta/n$, \rulehoef{} can certify
$\sqrt{\ln(1/\delta')/(2k)}\le\alpha$ only when
$k\ge\ln(1/\delta')/(2\alpha^2)$, whereas \rulecp{} needs only
$k\ge\ln(1/\delta')/\ln\tfrac{1}{1-\alpha}$, smaller by a factor of about
$2\alpha$. At $n{=}200$ and $\delta{=}0.1$ ($\delta'{=}5{\times}10^{-4}$)
the requirements are $380$ vs.\ $73$ accepted points at $\alpha{=}0.1$ and
$1520$ vs.\ $149$ at $\alpha{=}0.05$: on $200$ calibration points
\rulehoef{} cannot certify $\alpha{\le}0.1$ \emph{even for a perfect
detector}, while \rulecp{} succeeds as soon as a clean prefix of $73$
points exists. This single calculation predicts the pattern of zeros in
Table~\ref{tab:synth}.

\subsection{Exchangeability and a conservative mitigation}
\label{ssec:mitig-method}
Certificate~\eqref{eq:rule} requires calibration and test points to be
exchangeable. In grouped deployments (calibrate on machine types or
generators $\mathcal{G}_{\mathrm{cal}}$, test on disjoint
$\mathcal{G}_{\mathrm{te}}$) this premise fails. As a baseline mitigation we
compute a per-group threshold $\hat\lambda_g$ on each calibration group $g$
with budget split $\delta/|\mathcal{G}_{\mathrm{cal}}|$ and deploy the most
conservative one, $\hat\lambda_{\mathrm{mit}}=\min_g\hat\lambda_g$. This
guards against the calibration mixture hiding a hard group but still carries
no formal guarantee for unseen groups.

\section{Experiments}
\label{sec:exp}
All experiments are post hoc on frozen scores and CPU-only; $\delta{=}0.1$
and $\alpha\in\{0.05,0.1,0.2\}$ (plus $0.4$ for the weak real scores)
throughout. Violation rates are over independent repetitions (100 synthetic,
50 real); under exchangeability, certified rules are allowed up to
$\delta{=}10\%$. As a global negative control, pooling \emph{every}
certified-rule run under exchangeable random splits yields 0 violations in
2700 synthetic runs, 2 in 900 controlled-shift runs, and 2 in 3000 real-score
runs---so the violations reported below are not implementation artifacts.

\textbf{Data.} \emph{Synthetic}: $y\sim\mathrm{Bern}(\pi)$,
$s\mid y\sim\mathcal{N}(d'y,1)$ with $d'{=}1.5$, $\pi\in\{0.1,0.3,0.5\}$,
$n_{\mathrm{cal}}{=}200$, $n_{\mathrm{te}}{=}5000$.
\emph{ASD}: the DCASE~2023 Task~2 development set~\cite{dohi2023dcase},
built on ToyADMOS2~\cite{harada2021} and MIMII~DG~\cite{dohi2022mimii}
(7 machine types: ToyCar, ToyTrain, bearing, fan, gearbox, slider, valve;
200 clips each, balanced normal/anomalous, base rate $0.5$); we extract
BEATs embeddings~\cite{chen2023beats} and compute four standard anomaly-score
backends (two $k$-NN variants, Mahalanobis, PCA residual; test AUC
0.60--0.65). \emph{Forensics}: a GenImage~\cite{zhu2023genimage} subset with
800 real images and 7 generators (ADM, BigGAN, GLIDE, Midjourney, SD1.5,
VQDM, Wukong) $\times$ 100 fakes each, scored by a
VGG~\cite{simonyan2015} feature detector trained independently of
calibration (AUC $0.835$); although such detectors can transfer across GAN
architectures with careful augmentation~\cite{wang2020}, generalization to
unseen (e.g.\ diffusion-based) generators remains the known hard
case~\cite{ojha2023}, which is precisely
the grouped deployment we audit. Error events are
$y{=}1$: a missed anomaly (ASD) or a fake accepted as real (forensics).

\begin{table}[t]
\centering
\caption{Synthetic audit: mean coverage\,/\,violation rate over 100
repetitions ($n_{\mathrm{cal}}{=}200$, $n_{\mathrm{te}}{=}5000$,
$\delta{=}0.1$).}
\label{tab:synth}
\footnotesize
\setlength{\tabcolsep}{3.2pt}
\begin{tabular}{llccc}
\toprule
$\pi$ & Rule & $\alpha{=}0.05$ & $\alpha{=}0.1$ & $\alpha{=}0.2$ \\
\midrule
\multirow{4}{*}{0.1}
 & \rulehoef{}  & 0.000\,/\,0.00 & 0.000\,/\,0.00 & 0.762\,/\,0.00 \\
 & \rulecp{}    & 0.008\,/\,0.00 & \textbf{0.545}\,/\,0.00 & 0.987\,/\,0.00 \\
 & \rulewsr{}   & 0.008\,/\,0.00 & 0.494\,/\,0.00 & 0.969\,/\,0.00 \\
 & \rulenaive{} & 0.847\,/\,\textbf{0.49} & 0.982\,/\,0.09 & 0.994\,/\,0.00 \\
\midrule
\multirow{4}{*}{0.3}
 & \rulehoef{}  & 0.000\,/\,0.00 & 0.000\,/\,0.00 & 0.000\,/\,0.00 \\
 & \rulecp{}    & 0.000\,/\,0.00 & 0.000\,/\,0.00 & \textbf{0.481}\,/\,0.00 \\
 & \rulewsr{}   & 0.000\,/\,0.00 & 0.005\,/\,0.00 & 0.436\,/\,0.00 \\
 & \rulenaive{} & 0.401\,/\,\textbf{0.73} & 0.590\,/\,0.63 & 0.831\,/\,0.59 \\
\midrule
\multirow{4}{*}{0.5}
 & \rulehoef{}  & 0.000\,/\,0.00 & 0.000\,/\,0.00 & 0.000\,/\,0.00 \\
 & \rulecp{}    & 0.000\,/\,0.00 & 0.000\,/\,0.00 & 0.043\,/\,0.00 \\
 & \rulewsr{}   & 0.000\,/\,0.00 & 0.000\,/\,0.00 & 0.026\,/\,0.00 \\
 & \rulenaive{} & 0.168\,/\,\textbf{0.70} & 0.268\,/\,0.58 & 0.454\,/\,0.54 \\
\bottomrule
\end{tabular}
\end{table}

\begin{figure*}[t]
\centering
\includegraphics[width=0.92\textwidth]{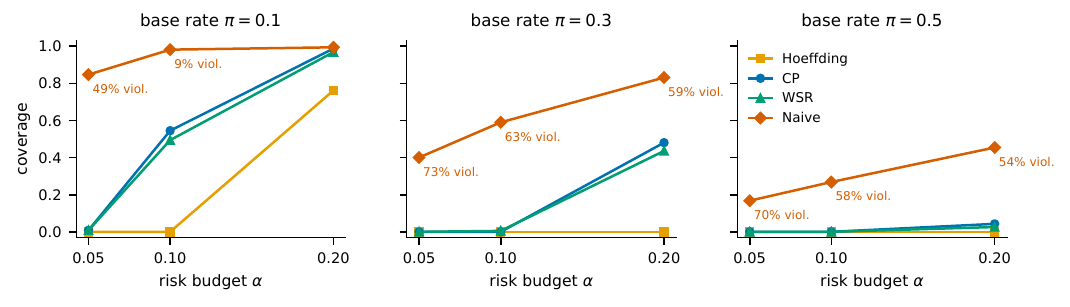}
\caption{Certified coverage vs.\ risk budget $\alpha$ on synthetic data
($n_{\mathrm{cal}}{=}200$, mean over 100 repetitions, $\delta{=}0.1$).
\rulenaive{} (orange diamonds) buys its high coverage with budget violations
(annotated rates); certified rules never violated in these runs, and
\rulecp{}/\rulewsr{} recover most of the coverage that \rulehoef{} forfeits.}
\label{fig:coverage}
\end{figure*}

\subsection{Synthetic audit: false safety and bound tightness}
\label{ssec:synth}
Table~\ref{tab:synth} and Fig.~\ref{fig:coverage} show the main audit.
\rulenaive{} attains high coverage but exceeds its declared budget in
49--73\% of trials at $\alpha{=}0.05$ across base rates---worst exactly
where a guarantee is needed most. \rulehoef{} is safe but certifies zero
coverage in 8 of 9 settings. \rulecp{} recovers substantial coverage at zero
observed violations ($0\!\to\!0.545$ at $\pi{=}0.1,\alpha{=}0.1$;
$0\!\to\!0.481$ at $\pi{=}0.3,\alpha{=}0.2$), and \rulewsr{} tracks it
closely ($0.494$, $0.436$). For binary losses \rulecp{} is exact, so
\rulewsr{}'s value lies in extending to bounded non-binary losses at
near-\rulecp{} tightness.

\subsection{Real scores under exchangeable splits}
\label{ssec:real}
Table~\ref{tab:real} evaluates the frozen real scores with random 50/50
calibration/test splits, where exchangeability holds by construction. The
false sense of safety reproduces: \rulenaive{} violates its budget in
20--68\% of ASD splits and 34--52\% of forensics splits. All certified rules
stay at 0--2\%, well within $\delta$. The price of certification depends on
score quality: on the strong forensics score, \rulecp{} certifies coverage
$0.103$ at $\alpha{=}0.2$ and $0.720$ at $\alpha{=}0.4$; on the weak ASD
scores (AUC ${\le}0.65$ at base rate $0.5$) every certified rule abstains
almost everywhere for $\alpha\le0.2$. We report this honest negative result
deliberately: with weak detectors there is nothing to certify, and only
\rulenaive{} pretends otherwise.

\begin{table}[t]
\centering
\caption{Real scores, random splits (exchangeable), 50 repetitions,
$\delta{=}0.1$: mean coverage\,/\,violation rate per backend (two of four
ASD backends shown; the others behave alike).}
\label{tab:real}
\footnotesize
\setlength{\tabcolsep}{1.8pt}
\begin{tabular}{llccc}
\toprule
Dataset & Rule & $\alpha{=}0.1$ & $\alpha{=}0.2$ & $\alpha{=}0.4$ \\
\midrule
\multirow{4}{*}{\shortstack[l]{Forensics\\(VGG, AUC .835)}}
 & \rulehoef{}  & 0.000\,/\,0.00 & 0.000\,/\,0.00 & 0.695\,/\,0.00 \\
 & \rulecp{}    & 0.000\,/\,0.00 & 0.103\,/\,0.00 & 0.720\,/\,0.00 \\
 & \rulewsr{}   & 0.000\,/\,0.00 & 0.037\,/\,0.00 & 0.707\,/\,0.00 \\
 & \rulenaive{} & 0.177\,/\,0.52 & 0.472\,/\,0.44 & 0.868\,/\,0.34 \\
\midrule
\multirow{4}{*}{\shortstack[l]{ASD $k$-NN cos\\(BEATs, AUC .637)}}
 & \rulehoef{}  & 0.000\,/\,0.00 & 0.000\,/\,0.00 & 0.000\,/\,0.00 \\
 & \rulecp{}    & 0.000\,/\,0.00 & 0.000\,/\,0.00 & 0.006\,/\,0.02 \\
 & \rulewsr{}   & 0.000\,/\,0.00 & 0.000\,/\,0.00 & 0.005\,/\,0.02 \\
 & \rulenaive{} & 0.009\,/\,0.42 & 0.033\,/\,0.68 & 0.497\,/\,0.58 \\
\midrule
\multirow{4}{*}{\shortstack[l]{ASD Mahal.\\(BEATs, AUC .651)}}
 & \rulehoef{}  & 0.000\,/\,0.00 & 0.000\,/\,0.00 & 0.000\,/\,0.00 \\
 & \rulecp{}    & 0.000\,/\,0.00 & 0.000\,/\,0.00 & 0.000\,/\,0.00 \\
 & \rulewsr{}   & 0.000\,/\,0.00 & 0.000\,/\,0.00 & 0.000\,/\,0.00 \\
 & \rulenaive{} & 0.013\,/\,0.20 & 0.022\,/\,0.58 & 0.603\,/\,0.62 \\
\bottomrule
\end{tabular}
\end{table}

\subsection{Exchangeability ablation: shift, not the bounds}
\label{ssec:shift}
\begin{figure*}[t]
\centering
\includegraphics[width=0.92\textwidth]{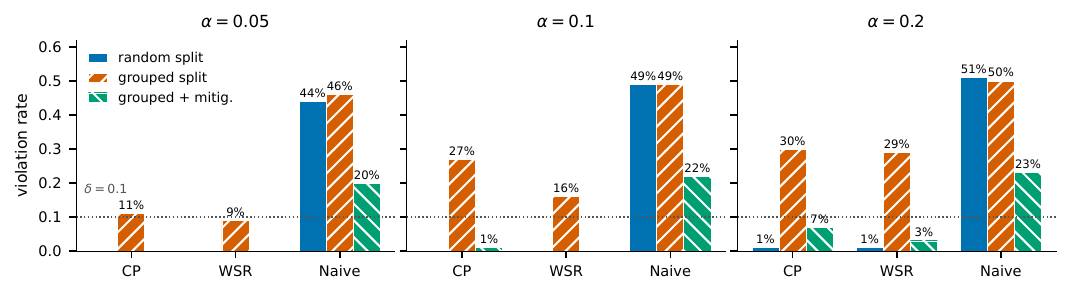}
\caption{Group shift, not the bounds, causes budget violations (controlled
synthetic shift, 100 repetitions; dotted line marks the tolerance
$\delta{=}0.1$ that certified rules are allowed under exchangeability).
Random splits keep \rulecp{}/\rulewsr{} at ${\le}1\%$; grouped splits push
the same rules to 9--30\%; the per-group conservative threshold (mitig.)
restores validity. \rulenaive{} overruns its budget throughout.}
\label{fig:shift}
\end{figure*}

We isolate the effect of grouped deployment with matched splits on the same
data. \emph{Controlled synthetic shift}: six groups with heterogeneous base
rates ($0.05$--$0.45$), separations ($d'{=}1.4$--$2.5$), and score offsets
($-0.4$--$0.6$); \texttt{random} mixes all groups into calibration and test,
\texttt{grouped} calibrates on 3 groups and tests on the other 3 (100
repetitions). Table~\ref{tab:shift} and Fig.~\ref{fig:shift}: under random
splits \rulecp{}/\rulewsr{} violate in at most 1\% of trials (within
$\delta{=}10\%$); under grouped splits the \emph{same rules on the same
data} violate in 9--30\%, exceeding $\delta$ at $\alpha\ge0.1$. Binomial
uncertainty does not explain this: the exact 95\% CI for the grouped
\rulecp{} rate is $[0.19,0.37]$ at $\alpha{=}0.1$ and $[0.21,0.40]$ at
$\alpha{=}0.2$, both excluding $\delta$. The failure
is therefore a property of the broken exchangeability premise, not of the
bounds. \rulenaive{} overruns in both conditions (44--51\%).

On the real scores the pattern recurs at practical budgets. Forensics
(generator holdout: calibrate on Real + 4 generators, test on Real + 3
unseen generators): \rulecp{} violations rise from 0\% (random) to 2\% at
$\alpha{=}0.1$ and 6\% at $\alpha{=}0.2$; \rulewsr{} to 2\%\,/\,4\%. ASD
(machine-type holdout, $\alpha{=}0.4$ where certified coverage is non-zero):
\rulecp{} rises from 0--2\% to 18--26\% on three of four backends, and
\rulewsr{} behaves similarly.

\begin{table}[t]
\centering
\caption{Controlled synthetic group shift, 100 repetitions, $\delta{=}0.1$:
violation rate by split protocol (mean coverage in parentheses for the
$\alpha{=}0.1$ rows).}
\label{tab:shift}
\footnotesize
\setlength{\tabcolsep}{2.8pt}
\begin{tabular}{lccc}
\toprule
Rule & Random & Grouped & Grouped$+$mitig. \\
\midrule
\rulecp{}, $\alpha{=}0.05$   & 0.00 & 0.11 & 0.00 \\
\rulecp{}, $\alpha{=}0.1$    & 0.00 (0.64) & \textbf{0.27} (0.62) & 0.01 (0.02) \\
\rulecp{}, $\alpha{=}0.2$    & 0.01 & 0.30 & 0.07 \\
\rulewsr{}, $\alpha{=}0.1$   & 0.00 (0.56) & \textbf{0.16} (0.53) & 0.00 (0.02) \\
\rulewsr{}, $\alpha{=}0.2$   & 0.01 & 0.29 & 0.03 \\
\rulenaive{}, $\alpha{=}0.1$ & 0.49 (0.80) & 0.49 (0.83) & 0.22 (0.61) \\
\bottomrule
\end{tabular}
\end{table}

\subsection{Failure mechanism: the hardest group breaks the budget}
\label{ssec:mech}
The grouped-split violations are not diffuse: they trace to identifiable
hard groups that the pooled calibration mixture averages away. On the
forensics score the seven generators are far from homogeneous---the
per-generator AUC against the shared real class spans $0.690$ (VQDM) to
$0.980$ (GLIDE)---and at the pooled \rulecp{} threshold for $\alpha{=}0.2$
the fake-acceptance rate among accepted images ranges from $0$--$2\%$
(BigGAN, GLIDE) to $23$--$30\%$ (Midjourney, VQDM). Whether a grouped run
violates is then almost a deterministic function of \emph{which} generators
are held out: over 100 grouped repetitions, \rulecp{} at $\alpha{=}0.2$
violated in $0\%$ of the runs whose test set contained at most one of the
three hardest generators (VQDM, ADM, SD1.5), in $21\%$ of those containing
two, and in $100\%$ of those containing all three. In ASD the anatomy
concentrates in one group: the weakest machine type (valve, per-machine
AUC $0.545$, $k$-NN backend) accounts for essentially all overruns---at
$\alpha{=}0.4$, \rulecp{} violated in $63\%$ of grouped runs with valve in
the test set and in $0\%$ of the rest. The general
reading: a pooled certificate is an \emph{on-average} statement over the
calibration mixture, and grouped deployment re-weights the test
distribution toward exactly the groups the certificate never audited. It
also explains why the controlled shift violates more often than the
forensics holdout: its groups differ in base rate and offset
simultaneously, the worst case for a pooled threshold.

\subsection{Mitigation trade-off}
\label{ssec:mitig}
The per-group conservative threshold of
Sec.~\ref{ssec:mitig-method} restores empirical validity: on the controlled
shift it pushes \rulecp{} violations from 11--30\% back to 0--7\% and
\rulewsr{} from 9--29\% to 0--3\% (Table~\ref{tab:shift}); on the forensics
holdout it removes all observed \rulecp{}/\rulewsr{} violations. The price
is severe: certified coverage collapses (\rulecp{} at $\alpha{=}0.1$:
$0.62\!\to\!0.02$; forensics at $\alpha{=}0.4$: $0.70\!\to\!0.03$), because
each group alone offers few calibration points and the minimum is dominated
by the hardest group. Two observations sharpen this trade-off. First, the
collapse is partly structural: each cell retains only $n_g{=}200$ of the
calibration points, so by the closed form of Sec.~2.2 \rulecp{} needs a
clean prefix of ${\approx}39$ accepted points per cell at $\alpha{=}0.2$,
which the hardest cell rarely offers; the minimum then certifies little
(mean coverage $0.90\!\to\!0.28$ at $\alpha{=}0.2$, now valid). Second,
the mitigation only repairs rules that had a certificate to begin with:
\rulenaive{} under the same per-group minimum still violated in 20--23\% of
the controlled-shift runs (Table~\ref{tab:shift}), so added conservatism is
no substitute for a bound. Together these quantify an open
robustness--utility gap that group-shift-aware certificates (e.g.\ weighted,
adaptive, robust, or beyond-exchangeability conformal
methods~\cite{tibshirani2019,gibbs2021,cauchois2024,barber2023}) would need
to close for selective risk on signals.

\section{Conclusion and limitations}
\label{sec:concl}
We audited selective-prediction risk certificates on signal-domain scores.
Uncertified empirical thresholding exceeds its declared budget in up to 73\%
of trials---a false sense of safety whenever it is reported as
``risk-controlled''. Exact and betting bounds (\rulecp{}, \rulewsr{}) are
the practical choices, recovering coverage that Hoeffding forfeits while
keeping violations within $\delta$ under exchangeability. Grouped
deployment breaks the premise of all certificates, and a conservative
per-group fix trades almost all coverage for validity. We recommend that
papers reporting risk-controlled selective systems state the bound type,
$\delta$, and the split protocol; report violation rates over repeated
splits with binomial confidence intervals; include a random-split negative
control to separate shift effects from implementation error; and release
frozen scores so the audit is reproducible. \emph{Limitations}: our losses
are binary, where \rulecp{} is already exact and \rulewsr{}'s advantage is
its generality to bounded losses; the real-data detectors are deliberately
simple frozen baselines, and stronger scores would enlarge the certified
regimes; the mitigation is a heuristic without guarantees for unseen
groups; and our group-shift evidence covers two signal tasks, not all
deployment regimes.

\bibliographystyle{IEEEbib}
\bibliography{refs}

\end{document}